\documentclass[a4paper]{styles/svproc}
\usepackage{url}
\usepackage{amsmath,amssymb,amsfonts}
\usepackage{algorithmic}
\usepackage{graphicx}
\usepackage{textcomp}
\usepackage{xcolor}
\usepackage{booktabs}
\usepackage{multirow} 
\usepackage{makecell}
\usepackage{xspace}

\usepackage{bm}

\newcommand{\arxivAdd}[1]{#1}
\newcommand{\arxivRemove}[1]{}

\begin{document}
\mainmatter              % start of a contribution
%
% \title{Object-Centric Scene Reconstruction via Hierarchical Superquadrics for Robust Map Alignment and Navigation}
% \title{Hierarchical Object Representation for Spatial Robot Perception}
% \title{Hierarchical Object Representation \\ for Spatial Robot Perception}
\title{Hierarchical Object Representation for Spatial Robot Perception: Points, Meshes, and Superquadrics}
% %
\titlerunning{Hierarchical Object Representation for Spatial Robot Perception}  % abbreviated title (for running head)
% %                                     also used for the TOC unless
% %                                     \toctitle is used
% %
\author{Ceng Zhang$^{1,2}$ \and Wan Su$^{1,2}$ \and  Mohamed Samshad$^{1}$ \and \\ Gregory S. Chirikjian$^{2}$ \and Rajat Talak$^{1}$}
% \author{Ceng Zhang\inst{1} \and Su Wan\inst{1} \and % Mohamed Samshad\inst{1} \and Rajat Talak\inst{1}}
%
\authorrunning{Ceng Zhang et al.} % abbreviated author list (for running head)
% \tocauthor{Ceng Zhang, Su Wan, Mohamed Samshad, Rajat Talak}
%
\institute{$^{1}$National University of Singapore (NUS)\\
% \email{zc\_ceng@u.nus.edu}, \email{suwan@u.nus.edu}, \email{samshad@nus.edu.sg}, \email{talak@nus.edu.sg} \\
$^{2}$Mohamed bin Zayed University of Artificial Intelligence (MBZUAI) \\
% \email{Ceng.Zhang@mbzuai.ac.ae}, \email{Wan.Su@mbzuai.ac.ae}, \email{Gregory.Chirikjian@mbzuai.ac.ae}
% \\ WWW home page: \texttt{http://users/\homedir iekeland/web/welcome.html}
}

\maketitle              % typeset the title of the contribution

%!TEX root = main.tex

% LC: can be inserted
% \newcommand{\qed}{{\hfill $\square$}}

%% Format definition
%\newrefformat{prob}{Problem\,\ref{#1}}
%\newrefformat{def}{Definition\,\ref{#1}}
%\newrefformat{sec}{Section\,\ref{#1}}
%\newrefformat{sub}{Section\,\ref{#1}}
%\newrefformat{prop}{Proposition\,\ref{#1}}
%\newrefformat{app}{Appendix\,\ref{#1}}
%\newrefformat{alg}{Algorithm\,\ref{#1}}
%\newrefformat{cor}{Corollary\,\ref{#1}}
%\newrefformat{thm}{Theorem\,\ref{#1}}
%\newrefformat{lem}{Lemma\,\ref{#1}}
%\newrefformat{fig}{Fig.\,\ref{#1}}
%\newrefformat{tab}{Table\,\ref{#1}}

% Problem environment
% \newtheorem{theorem}{Theorem}
% \newtheorem{problem}{Problem}
% \newtheorem{trule}[theorem]{Rule}
% \newtheorem{corollary}[theorem]{Corollary}
%\newtheorem{algorithm}{Algorithm}
%\newtheorem{procedure}{\textbf{Procedure}}
% \newtheorem{conjecture}[theorem]{Conjecture}
% \newtheorem{lemma}[theorem]{Lemma}
% \newtheorem{assumption}[theorem]{Assumption}
% \newtheorem{definition}[theorem]{Definition}
% \newtheorem{proposition}[theorem]{Proposition}
% \newtheorem{remark}[theorem]{Remark}
% \newtheorem{example}[theorem]{Example}

% Shortcuts
\newcommand{\cf}{\emph{cf.}\xspace}
\newcommand{\Figure}{Fig.}
\newcommand{\bdmath}{\begin{dmath}}
\newcommand{\edmath}{\end{dmath}}
\newcommand{\beq}{\begin{equation}}
\newcommand{\eeq}{\end{equation}}
\newcommand{\bdm}{\begin{displaymath}}
\newcommand{\edm}{\end{displaymath}}
\newcommand{\bea}{\begin{eqnarray}}
\newcommand{\eea}{\end{eqnarray}}
\newcommand{\beal}{\beq \begin{array}{ll}}
\newcommand{\eeal}{\end{array} \eeq}
\newcommand{\beas}{\begin{eqnarray*}}
\newcommand{\eeas}{\end{eqnarray*}}
\newcommand{\ba}{\begin{array}}
\newcommand{\ea}{\end{array}}
\newcommand{\bit}{\begin{itemize}}
\newcommand{\eit}{\end{itemize}}
\newcommand{\ben}{\begin{enumerate}}
\newcommand{\een}{\end{enumerate}}
\newcommand{\expl}[1]{&&\qquad\text{\color{gray}(#1)}\nonumber}

\newcommand{\newDay}[1]{
	\noindent\rule{\linewidth}{1pt}
	\texttt{Date: #1} \\
	\rule{\linewidth}{1pt}
}

% Calligraphic fonts
\newcommand{\calA}{{\cal A}}
\newcommand{\calB}{{\cal B}}
\newcommand{\calC}{{\cal C}}
\newcommand{\calD}{{\cal D}}
\newcommand{\calE}{{\cal E}}
\newcommand{\calF}{{\cal F}}
\newcommand{\calG}{{\cal G}}
\newcommand{\calH}{{\cal H}}
\newcommand{\calI}{{\cal I}}
\newcommand{\calJ}{{\cal J}}
\newcommand{\calK}{{\cal K}}
\newcommand{\calL}{{\cal L}}
\newcommand{\calM}{{\cal M}}
\newcommand{\calN}{{\cal N}}
\newcommand{\calO}{{\cal O}}
\newcommand{\calP}{{\cal P}}
\newcommand{\calQ}{{\cal Q}}
\newcommand{\calR}{{\cal R}}
\newcommand{\calS}{{\cal S}}
\newcommand{\calT}{{\cal T}}
\newcommand{\calU}{{\cal U}}
\newcommand{\calV}{{\cal V}}
\newcommand{\calW}{{\cal W}}
\newcommand{\calX}{{\cal X}}
\newcommand{\calY}{{\cal Y}}
\newcommand{\calZ}{{\cal Z}}

% SETS:
\newcommand{\setA}{\textsf{A}}
\newcommand{\setB}{\textsf{B}}
\newcommand{\setC}{\textsf{C}}
\newcommand{\setD}{\textsf{D}}
\newcommand{\setE}{\textsf{E}}
\newcommand{\setF}{\textsf{F}}
\newcommand{\setG}{\textsf{G}}
\newcommand{\setH}{\textsf{H}}
\newcommand{\setI}{\textsf{I}}
\newcommand{\setJ}{\textsf{J}}
\newcommand{\setK}{\textsf{K}}
\newcommand{\setL}{\textsf{L}}
\newcommand{\setM}{\textsf{M}}
\newcommand{\setN}{\textsf{N}}
\newcommand{\setO}{\textsf{O}}
\newcommand{\setP}{\textsf{P}}
\newcommand{\setQ}{\textsf{Q}}
\newcommand{\setR}{\textsf{R}}
\newcommand{\setS}{\textsf{S}}
\newcommand{\setT}{\textsf{T}}
\newcommand{\setU}{\textsf{U}}
\newcommand{\setV}{\textsf{V}}
\newcommand{\setW}{\textsf{W}}
\newcommand{\setX}{\textsf{X}}
\newcommand{\setY}{\textsf{Y}}
\newcommand{\setZ}{\textsf{Z}}

%General
\newcommand{\smallheading}[1]{\textit{#1}: }
\newcommand{\algostep}[1]{{\small\texttt{#1:}}\xspace}
\newcommand{\etal}{\emph{et~al.}\xspace}
\newcommand{\setal}{~\emph{et~al.}\xspace}
\newcommand{\eg}{\emph{e.g.,}\xspace}
\newcommand{\ie}{\emph{i.e.,}\xspace}
\newcommand{\myParagraph}[1]{{\bf #1.}\xspace}
% \newcommand{\email}[1]{{\smaller \textsf{#1}}}

%Typography
\newcommand{\M}[1]{{\bm #1}} % Face for matrices
\renewcommand{\boldsymbol}[1]{{\bm #1}}
\newcommand{\algoname}[1]{\textsc{#1}} % Name of algorithms

%Editing
\newcommand{\towrite}{{\color{red} To write}\xspace}
\newcommand{\xxx}{{\color{red} XXX}\xspace}
\newcommand{\AC}[1]{{\color{red} \textbf{AC}: #1}}
\newcommand{\LC}[1]{{\color{red} \textbf{LC}: #1}}
\newcommand{\FM}[1]{{\color{red} \textbf{FM}: #1}}
\newcommand{\RT}[1]{{\color{blue} \textbf{RT}: #1}}
\newcommand{\VT}[1]{{\color{blue} \textbf{VT}: #1}}
\newcommand{\FK}[1]{{\color{violet} \textbf{FK}: #1}}
\newcommand{\hide}[1]{}
\newcommand{\wrt}{w.r.t.\xspace}
\newcommand{\highlight}[1]{{\color{red} #1}}
\newcommand{\tocheck}[1]{{\color{brown} #1}}
\newcommand{\grayout}[1]{{\color{gray} #1}}
\newcommand{\grayText}[1]{{\color{gray} \text{#1} }}
\newcommand{\grayMath}[1]{{\color{gray} #1 }}
\newcommand{\hiddenText}{{\color{gray} hidden text.}}
\newcommand{\hideWithText}[1]{\hiddenText}

\newcommand{\NA}{{\sf n/a}}
\newcommand{\versus}{\scenario{VS}\xspace}

%Basic math symbols
\newcommand{\kron}{\otimes}
\newcommand{\dist}{\mathbf{dist}}
\newcommand{\iter}{\! \rm{iter.} \;}
\newcommand{\leqt}{\!\!\! < \!\!\!}
\newcommand{\geqt}{\!\!\! > \!\!\!}
\newcommand{\mysetminus}{-} % One set minus another
\newcommand{\powerset}{\mathcal{P}}
\newcommand{\Int}[1]{ { {\mathbb Z}^{#1} } }
\newcommand{\Natural}[1]{ { {\mathbb N}^{#1} } }
\newcommand{\Complex}[1]{ { {\mathbb C}^{#1} } }
\newcommand{\one}{ {\mathbf{1}} }
\newcommand{\subject}{\text{ subject to }}
% \DeclareMathOperator*{\argmax}{arg\,max}
% \DeclareMathOperator*{\argmin}{arg\,min}

%% Norms
\newcommand{\normsq}[2]{\left\|#1\right\|^2_{#2}}
\newcommand{\norm}[1]{\left\| #1 \right\|}
\newcommand{\normsqs}[2]{\|#1\|^2_{#2}}
\newcommand{\infnorm}[1]{\left\|#1\right\|_{\infty}}
\newcommand{\zeronorm}[1]{\|#1\|_{0}}
\newcommand{\onenorm}[1]{\|#1\|_{1}}
\newcommand{\lzero}{\ell_{0}}
\newcommand{\lone}{\ell_{1}}
\newcommand{\linf}{\ell_{\infty}}

\newcommand{\E}{{\mathbb{E}}}
\newcommand{\EV}{\mathbb{E}}
\newcommand{\erf}{{\mathbf{erf}}}
\newcommand{\prob}[1]{{\mathbb P}\left(#1\right)}
\newcommand{\tran}{^{\mathsf{T}}}
\newcommand{\traninv}{^{-\mathsf{T}}}
\newcommand{\diag}[1]{\mathrm{diag}\left(#1\right)}
\newcommand{\trace}[1]{\mathrm{tr}\left(#1\right)}
\newcommand{\conv}[1]{\mathrm{conv}\left(#1\right)}
\newcommand{\polar}[1]{\mathrm{polar}\left(#1\right)}
\newcommand{\rank}[1]{\mathrm{rank}\left(#1\right)}
\newcommand{\e}{{\mathrm e}}
\newcommand{\inv}{^{-1}}
\newcommand{\pinv}{^\dag}
\newcommand{\until}[1]{\{1,\dots, #1\}}
\newcommand{\ones}{{\mathbf 1}}
\newcommand{\zero}{{\mathbf 0}}
\newcommand{\eye}{{\mathbf I}}
\newcommand{\vect}[1]{\left[\begin{array}{c}  #1  \end{array}\right]}
\newcommand{\matTwo}[1]{\left[\begin{array}{cc}  #1  \end{array}\right]}
\newcommand{\matThree}[1]{\left[\begin{array}{ccc}  #1  \end{array}\right]}
\newcommand{\dss}{\displaystyle}
\newcommand{\Real}[1]{ { {\mathbb R}^{#1} } }
\newcommand{\reals}{\Real{}}
\newcommand{\opt}{^{\star}}
\newcommand{\only}{^{\alpha}}
\newcommand{\copt}{^{\text{c}\star}}
\newcommand{\atk}{^{(k)}}
\newcommand{\att}{^{(t)}}
\newcommand{\at}[1]{^{(#1)}}
\newcommand{\attau}{^{(\tau)}}
\newcommand{\atc}[1]{^{(#1)}}
\newcommand{\atK}{^{(K)}}
\newcommand{\atj}{^{(j)}}
\newcommand{\projector}{{\tt projector}}
\newcommand{\setdef}[2]{ \{#1 \; {:} \; #2 \} }
\newcommand{\smalleye}{\left(\begin{smallmatrix}1&0\\0&1\end{smallmatrix}\right)}

%Spaces
\newcommand{\SEtwo}{\ensuremath{\mathrm{SE}(2)}\xspace}
\newcommand{\SE}[1]{\ensuremath{\mathrm{SE}(#1)}\xspace}
\newcommand{\SEthree}{\ensuremath{\mathrm{SE}(3)}\xspace}
\newcommand{\SOtwo}{\ensuremath{\mathrm{SO}(2)}\xspace}
\newcommand{\SOthree}{\ensuremath{\mathrm{SO}(3)}\xspace}
\newcommand{\Othree}{\ensuremath{\mathrm{O}(3)}\xspace}
\newcommand{\SOn}{\ensuremath{\mathrm{SO}(n)}\xspace}
\newcommand{\SO}[1]{\ensuremath{\mathrm{SO}(#1)}\xspace}
\newcommand{\On}{\ensuremath{\mathrm{O}(n)}\xspace}
\newcommand{\sotwo}{\ensuremath{\mathrm{so}(2)}\xspace}
\newcommand{\sothree}{\ensuremath{\mathrm{so}(3)}\xspace}
\newcommand{\intexpmap}[1]{\mathrm{Exp}\left(#1\right)}
\newcommand{\intlogmap}[1]{\mathrm{Log}\left(#1\right)}
\newcommand{\logmapz}[1]{\mathrm{Log}_0(#1)}
\newcommand{\loglikelihood}{\mathrm{log}\calL}
\newcommand{\intprinlogmap}{\mathrm{Log}_0}
\newcommand{\expmap}[1]{\intexpmap{#1}}
\newcommand{\expmaps}[1]{\langle #1 \rangle_{2\pi}}
\newcommand{\biggexpmap}[1]{\left\langle #1 \right\rangle_{2\pi}}
\newcommand{\logmap}[1]{\intlogmap{#1}}
\newcommand{\round}[1]{\mathrm{round}\left( #1 \right)}
\newcommand{\Rthree}{\ensuremath{\mathbb{R}^3}\xspace}
\newcommand{\Rtwo}{\ensuremath{\mathbb{R}^{2}}\xspace}
\newcommand{\R}[1]{\ensuremath{\mathbb{R}^{#1}}\xspace}

% Matrices
\newcommand{\MA}{\M{A}}
\newcommand{\MB}{\M{B}}
\newcommand{\MC}{\M{C}}
\newcommand{\MD}{\M{D}}
\newcommand{\ME}{\M{E}}
\newcommand{\MJ}{\M{J}}
\newcommand{\MK}{\M{K}}
\newcommand{\MG}{\M{G}}
\newcommand{\MM}{\M{M}}
\newcommand{\MN}{\M{N}}
\newcommand{\MP}{\M{P}}
\newcommand{\MQ}{\M{Q}}
\newcommand{\MU}{\M{U}}
\newcommand{\MR}{\M{R}}
\newcommand{\MS}{\M{S}}
\newcommand{\MI}{\M{I}}
\newcommand{\MV}{\M{V}}
\newcommand{\MF}{\M{F}}
\newcommand{\MH}{\M{H}}
\newcommand{\ML}{\M{L}}
\newcommand{\MO}{\M{O}}
\newcommand{\MT}{\M{T}}
\newcommand{\MX}{\M{X}}
\newcommand{\MY}{\M{Y}}
\newcommand{\MW}{\M{W}}
\newcommand{\MZ}{\M{Z}}
\newcommand{\MSigma}{\M{\Sigma}}
\newcommand{\MOmega}{\M{\Omega}}
\newcommand{\MPhi}{\M{\Phi}}
\newcommand{\MPsi}{\M{\Psi}}
\newcommand{\MDelta}{\M{\Delta}}
\newcommand{\MLambda}{\M{\Lambda}}

% vectors
\newcommand{\vzero}{\boldsymbol{0}}
\newcommand{\vone}{\boldsymbol{1}}
\newcommand{\va}{\boldsymbol{a}}
\newcommand{\vh}{\boldsymbol{h}}
\newcommand{\vb}{\boldsymbol{b}}
\newcommand{\vc}{\boldsymbol{c}}
\newcommand{\vd}{\boldsymbol{d}}
\newcommand{\ve}{\boldsymbol{e}}
\newcommand{\vf}{\boldsymbol{f}}
\newcommand{\vg}{\boldsymbol{g}}
\newcommand{\vk}{\boldsymbol{k}}
\newcommand{\vl}{\boldsymbol{l}}
\newcommand{\vn}{\boldsymbol{n}}
\newcommand{\vo}{\boldsymbol{o}}
\newcommand{\vp}{\boldsymbol{p}}
\newcommand{\vq}{\boldsymbol{q}}
\newcommand{\vr}{\boldsymbol{r}}
\newcommand{\vs}{\boldsymbol{s}}
\newcommand{\vu}{\boldsymbol{u}}
\newcommand{\vv}{\boldsymbol{v}}
\newcommand{\vt}{\boldsymbol{t}}
\newcommand{\vxx}{\boldsymbol{x}}
\newcommand{\vy}{\boldsymbol{y}}
\newcommand{\vw}{\boldsymbol{w}}
\newcommand{\vzz}{\boldsymbol{z}}
\newcommand{\vdelta}{\boldsymbol{\delta}}
\newcommand{\vgamma}{\boldsymbol{\gamma}}
\newcommand{\vlambda}{\boldsymbol{\lambda}}
\newcommand{\vtheta}{\boldsymbol{\theta}}
\newcommand{\valpha}{\boldsymbol{\alpha}}
\newcommand{\vbeta}{\boldsymbol{\beta}}
\newcommand{\vnu}{\boldsymbol{\nu}}
\newcommand{\vmu}{\boldsymbol{\mu}}
\newcommand{\vepsilon}{\boldsymbol{\epsilon}}
\newcommand{\vtau}{\boldsymbol{\tau}}

%Intrinsic geometry
\newcommand{\Rtheta}{\boldsymbol{R}}
\newcommand{\symf}{f} % Symmetry function

%Angles
\newcommand{\angledomain}{(-\pi,+\pi]}

% Tree, graphs, and cycle basis
\newcommand{\MCB}{\mathsf{MCB}}
\newcommand{\FCM}{\mathsf{FCM}}
\newcommand{\FCB}{\mathsf{FCB}}
\newcommand{\cyclemap}[1]{\calC^{\calG}\left(#1\right)}
\newcommand{\incidencemap}[1]{\calA^{\calG}\left(#1\right)}
\newcommand{\cyclemapk}{\calC^{\calG}_{k}}
\newcommand{\incidencemapij}{\calA^{\calG}_{ij}}
\renewcommand{\ij}{_{ij}}
\newcommand{\foralledges}{\forall(i,j) \in \calE}
\newcommand{\sumalledges}{
     \displaystyle
     \sum_{(i,j) \in \calE}}
\newcommand{\sumalledgesm}{
     \displaystyle
     \sum_{i=1}^{m}}
\newcommand{\T}{\mathsf{T}}
\newcommand{\To}{\T_{\rm o}}
\newcommand{\Tm}{\T_{\rm m}}
\newcommand{\MCBa}{\MCB_{\mathsf{a}}}
\newcommand{\FCBo}{\FCB_{\mathsf{o}}}
\newcommand{\FCBm}{\FCB_{\mathsf{m}}}

% Algorithms
\newcommand{\algoonlyname}{MOLE2D}
\newcommand{\algoml}{{\smaller\sf \algoonlyname}\xspace}
\newcommand{\algocyclebasis}{\algoname{compute-cycle-basis}}
\newcommand{\scenario}[1]{{\smaller \sf#1}\xspace}
\newcommand{\toro}{{\smaller\sf Toro}\xspace}
\newcommand{\gtwoo}{{\smaller\sf g2o}\xspace}
\newcommand{\gtwooST}{{\smaller\sf g2oST}\xspace}
\newcommand{\gtwood}{{\smaller\sf g2o{10}}\xspace}
\newcommand{\gtsam}{{\smaller\sf gtsam}\xspace}
\newcommand{\isam}{{\smaller\sf iSAM}\xspace}
\newcommand{\lago}{{\smaller\sf LAGO}\xspace}
\newcommand{\egtwoo}{{\smaller\sf \algoonlyname+g2o}\xspace}

% Datasets
%\newcommand{\grid}{\scenario{cube}}
\newcommand{\rim}{\scenario{rim}}
\newcommand{\cubicle}{\scenario{cubicle}}
\newcommand{\sphere}{\scenario{sphere}}
\newcommand{\sphereHard}{\scenario{sphere-a}}
\newcommand{\garage}{\scenario{garage}}
\newcommand{\torus}{\scenario{torus}}
\newcommand{\oneloop}{\scenario{circle}}
\newcommand{\intel}{\scenario{INTEL}}
\newcommand{\bovisa}{\scenario{Bovisa}}
\newcommand{\bov}{\scenario{B25b}}
\newcommand{\fra}{\scenario{FR079}}
\newcommand{\frb}{\scenario{FRH}}
\newcommand{\csail}{\scenario{CSAIL}}
\newcommand{\Ma}{\scenario{M3500}}
\newcommand{\Mb}{\scenario{M10000}}
\newcommand{\ATE}{\scenario{ATE}}
\newcommand{\CVX}{\scenario{CVX}}
\newcommand{\NEOS}{\scenario{NEOS}}
\newcommand{\sdptThree}{\scenario{sdpt3}}
\newcommand{\MOSEK}{\scenario{MOSEK}}
\newcommand{\NESTA}{\scenario{NESTA}}
\newcommand{\vertigo}{\scenario{Vertigo}}
\newcommand{\SDPA}{\scenario{SDPA}}

\newcommand{\cvx}{{\sf cvx}\xspace}

% COLORS
\newcommand{\blue}[1]{{\color{blue}#1}}
\newcommand{\green}[1]{{\color{green}#1}}
\newcommand{\red}[1]{{\color{red}#1}}

% TO MANAGE REFERENCES
%============================================================================
\newcommand{\linkToPdf}[1]{\href{#1}{\blue{(pdf)}}}
\newcommand{\linkToPpt}[1]{\href{#1}{\blue{(ppt)}}}
\newcommand{\linkToCode}[1]{\href{#1}{\blue{(code)}}}
\newcommand{\linkToWeb}[1]{\href{#1}{\blue{(web)}}}
\newcommand{\linkToVideo}[1]{\href{#1}{\blue{(video)}}}
\newcommand{\linkToMedia}[1]{\href{#1}{\blue{(media)}}}
\newcommand{\award}[1]{\xspace} % {{\red{#1}}} % omit awards

% PAPER-SPECIFIC COMMANDS
%============================================================================

% Linear approximation
\newcommand{\vpose}{\boldsymbol{x}}
\newcommand{\vz}{\boldsymbol{z}}
\newcommand{\vDelta}{\boldsymbol{\Delta}}
\newcommand{\vposesub}{\hat{\vpose}}
\newcommand{\vpossub}{\hat{\vpos}}
\newcommand{\vthetasub}{\hat{\vtheta}}
\newcommand{\Pthetasub}{\MP_\vtheta}
\newcommand{\thetasub}{\hat{\theta}}
\newcommand{\vposecorr}{\tilde{\vpose}}
\newcommand{\vposcorr}{\tilde{\vpos}}
\newcommand{\vthetacorr}{\tilde{\vtheta}}
\newcommand{\vposestar}{{\vpose}^{\star}}
\newcommand{\vposstar}{{\vpos}^{\star}}
\newcommand{\vthetastar}{{\vtheta}^{\star}}
\newcommand{\vcthetastar}{{\vctheta}^{\star}}
\newcommand{\pose}{\boldsymbol{x}}
\newcommand{\pos}{\boldsymbol{p}}
\newcommand{\mease}{z} % element
\newcommand{\meas}{\boldsymbol{z}} % vector
\newcommand{\meashat}{\hat{\meas}}

\newcommand{\threeDSG}{\text{3DSG}\xspace}
\newcommand{\hickory}{\texttt{Hickory}\xspace}
\begin{abstract}
% Github: hickory 
Hierarchical 3D Scene Graphs (\threeDSG) have emerged as an actionable and scalable representation for long-term autonomy incorporating metric, semantic, and topological information in the scene.
However, the question of geometric representation of objects in \threeDSG has been overlooked as most methods use simplified geometric models such as partial point clouds or 3D bounding boxes. 
In this work, we introduce a hierarchical object representation 
that can be leveraged for 
high-fidelity object-level reconstruction, 
object-based robust re-localization or map alignment,
and
efficient and analytical collision checking for safe robot navigation planning in dense and cluttered environments.
The representation is structurally organized into four distinct layers, progressively abstracting the scene from raw sensor data to dense 3D meshes to analytical primitives such as superquadrics, which provide a sparse and analytical representation for object geometry. 
We develop a pipeline that builds the hierarchical object representation from RGB-D image stream captured by a robot, and demonstrate its working in real-world open-set object scenes in both indoor and outdoor environments. 
%
% Our pipeline leverages the recent progress in object reconstruction 
%
Extensive experiments across diverse datasets including HOPE, ReplicaCAD, Kimera-Multi, and NUS Campus Dataset collected using Unitree B2 Robot validate our pipeline in both indoor and outdoor environments.  
We show that our superquadric-based map alignment method outperforms the current state-of-the-art object-based map alignment method ROMAN.
Our code can be found at https://github.com/perceptica-robotics/Hickory.

\end{abstract}
\vspace{-5mm}
\begin{figure}[t]
\centering
\includegraphics[width=\linewidth]{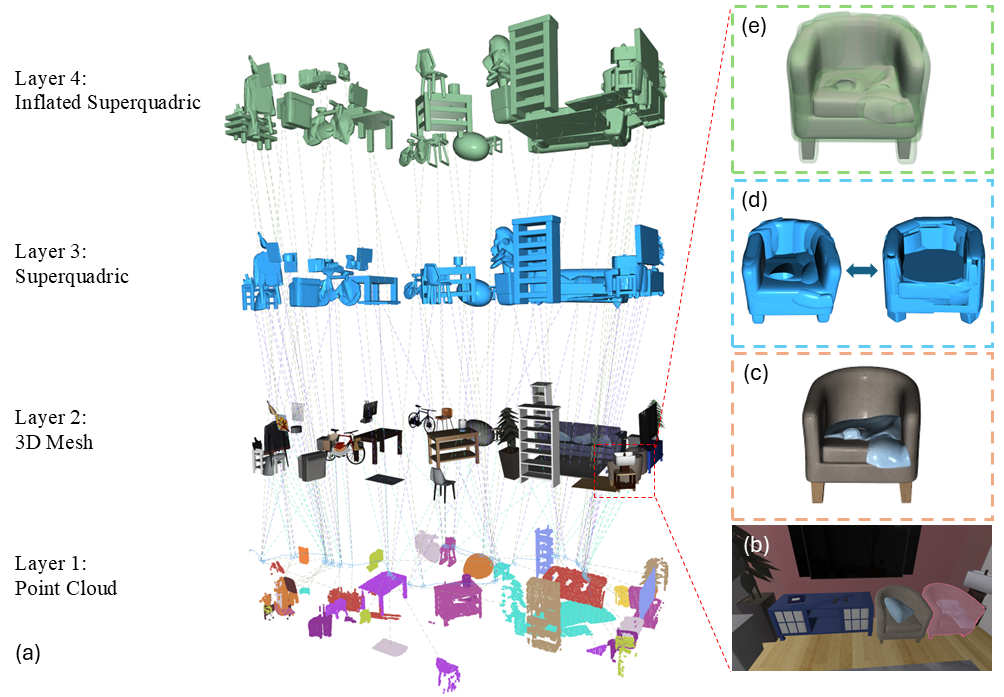}
\arxivRemove{\vspace{-9mm}}
\arxivAdd{\vspace{-5mm}}
\caption{Hierarchical Object Representation. (a) Our object representation is composed of four layers: Layer 1 (point cloud), Layer 2 (dense 3D mesh), Layer 3 (Superquadric), and Layer 4 (Inflated Superquadric). (b) Layer 1 (point cloud) is obtained by aggregating object point cloud trackied from multiple RGB-D views; (c) Layer 2 (dense 3D mesh) is obtained from a vision foundation model trained to generate 3D meshes from RGB-D views; (d) Layer 3 (Superquadrics) is extracted from Layer 2 and serves as a sparse parametric representation for object-based re-localization and map alignment; (e) Layer 4 (Inflated Superquadrics) is extracted from Layer 3 and serves as an efficient object representation for collision checking and navigation.}
\label{overview}
\arxivRemove{\vspace{-6mm}}
\arxivAdd{\vspace{-4mm}}
\end{figure}

\section{Introduction}
\label{sec:intro}
Spatial scene representation is fundamental problem for autonomous robots in novel and complex environments --- a good representation is meant to make downstream tasks (\eg navigation, manipulation, multi-robot localization) easier. Hierarchical 3D Scene Graphs (\threeDSG) have emerged as a powerful representation paradigm, elevating traditional metric and metric-semantic maps by encoding semantic, topological, and geometric relationships into a structured format \cite{rosinol20203d}. While much attention has been dedicated to modeling object-semantic, topological, inter-object relationships, the existing \threeDSG pipelines represent objects as overly simplified geometric models, such as partial point clouds or 3D bounding boxes~\cite{hughes2024foundations,wang2021dsp,catalano20253d,mascaro2025scene}. \emph{The question of geometric representation of the objects themselves has been overlooked.} 
% the geometric representation of the object nodes themselves has been overlooked.
%
% Existing \threeDSG pipelines represent objects as overly simplified geometric models, such as partial point clouds or 3D bounding boxes \cite{hughes2022hydra} \cite{wang2021dsp}. %While computationally efficient, these coarse representations fail to capture true geometric fidelity and dense topology, and very few works have achieved high-fidelity object-centric reconstruction in open-set environments.

The advent of vision foundation models offers a promising solution to this bottleneck. Recent object reconstruction models, such as LRM \cite{hong2024lrm}, InstantMesh \cite{xu2024instantmesh}, TRELLIS \cite{xiang2025structured} and SAM3D \cite{chen2025sam}, have demonstrated strong capabilities in recovering 3D object geometry from visual observations. 
% (RT -- It's ok. Let's cancel the justification. It appears weak.) Among them, we adopt SAM3D because it is well suited for visually grounded object reconstruction in cluttered scenes and can be naturally integrated with our object-centric RGB-D pipeline for metric-scale scene reconstruction.
% We use SAM3D because its object-centric reconstruction from masked RGB inputs naturally fits our segmentation-based RGB-D pipeline. 
These models enable high-fidelity, object-level reconstruction directly from a single RGB-D image, extracting complete geometric structures --- a vital prerequisite for downstream robot planning, especially in dense and cluttered environments.

However, integrating these dense, heavy mesh models directly into a \threeDSG creates a severe computational burden, particularly for onboard robotic applications. Utilizing raw, high-poly meshes for downstream tasks, such as collision avoidance or multi-session map alignment, is computationally prohibitive. This fundamental conflict between geometric fidelity and computational efficiency motivates the critical need for novel hierarchical object representation --- one that can abstract high-fidelity meshes into mathematically compact, expressive primitives without losing essential structural integrity.

% To address these challenges, 
Our \textbf{first} contribution is that 
we propose a novel \emph{hierarchical object representation} to address these challenges. See Figure~\ref{overview}.  
%that integrates high-fidelity perception with compact geometric abstraction. 
Our representation is structurally organized into four distinct layers, progressively abstracting the scene from raw sensor data to dense 3D meshes to analytical primitives. 
Layer 1 maintains the raw object point cloud, which are continuously captured and accumulated by our mapping and object registration module. Building upon this, Layer 2 consists of reconstructed dense object 3D meshes and Layer 3 stores a superquadric representation of the object~\cite{paschalidou2019superquadrics}. Layer 3 is specifically designed for long-term object-based robust re-localization and map alignment. Layer 4 stores inflated superquadrics designed for efficient collision checking and safe robot navigation (by allowing a safety distance) in dense and cluttered scenes. 

Our \textbf{second} contribution is that we develop a pipeline that builds the hierarchical object representation from streaming posed RGB-D images. Our pipeline works in real-world open-set object scenes in both indoor and outdoor environments. Our pipeline utilize SAM3D~\cite{chen2025sam} combined with FastSAM~\cite{zhao2023fast} for open-set segmentation, and introduce a novel semantic-driven best-view selection strategy for each object to ensure accurate geometric reconstruction from single-view images. Our pipeline leverages existing superquadric fitting~\cite{liu2023marching}, and propose a novel object association method by directly comparing superquadric parameters.
Our pipeline introduces a novel procedure for building Inflated Superquadrics and formulates a closed-form collision-checking mechanism using them. This reduces the computation of distance queries and enabling fast and mathematically guaranteed collision-free navigation.

Our \textbf{third} contribution is an extensive experimental evaluation across simulated and real-world scenes. We show that the proposed pipeline enables (i) high-fidelity object-level reconstruction through feature-guided best-view selection, (ii) robust object-based map alignment across different sequences using superquadric-based data association, and (iii) efficient collision-free navigation in dense and cluttered environments using inflated superquadric representations.
We test our pipeline on synthetic datasets such as ReplicaCAD~\cite{szot2021habitat}, real-world datasets such as HOPE~\cite{tyree2022hope}, Kimera-Multi~\cite{tian2022kimera}, and our own NUS Campus dataset collected on-board Unitree B2 Robot. We show best performance in comparison to many baselines in these experiments. In particular, our superquadric-based map alignment method outperforms the current state-of-the-art object-based map alignment method ROMAN~\cite{peterson2025roman}.

\section{Related Works}
\arxivAdd{Object-centric representations provide compact and semantic abstractions of environments, supporting object-level mapping, global localization, and structured scene understanding. %In this section, 
We review related work along these directions.}

\myParagraph{Object representation} 
Traditional SLAM systems rely on low-level features or photometric information \cite{mur2017orb,engel2014lsd}, which lack semantic structure and are sensitive to environmental changes \cite{tian2023resilient}. Object-level representations instead represent scenes as meaningful entities, improving robustness to viewpoint, illumination, and appearance variations.
Early methods represent objects using simple primitives such as points, cuboids, or quadrics. QuadricSLAM \cite{nicholson2018quadricslam} models objects as dual quadrics, while Mishima et al. \cite{mishima2018rgb} detect cuboids from planar structures. These representations are efficient but cannot capture complex geometries. Higher-fidelity methods, such as Fusion++ \cite{mccormac2018fusion++} and MaskFusion \cite{runz2018maskfusion}, use TSDF volumes or surfels for object reconstruction.
Recent foundation models enable open-set object perception and detailed reconstruction without predefined categories. SAM \cite{kirillov2023segment} and SAM3D \cite{chen2025sam} provide strong generalization for segmentation and mesh generation. In parallel, superquadrics offer compact and analytical object representations, supporting efficient pose estimation and shape abstraction in object-level SLAM \cite{han2023sq,tschopp2021superquadric}.

\myParagraph{Hierarchical 3D scene graphs}
3D scene graphs integrate geometry, semantics, and spatial relationships into structured scene representations. Armeni et al. \cite{armeni20193d} introduced a hierarchical graph with building, room, object, and camera-pose layers, while Kimera-DSG \cite{rosinol2020kimera} extended this formulation to dynamic scene graphs with both metric-semantic and topological layers.
Subsequent systems such as Hydra \cite{hughes2022hydra}, Hydra-Multi \cite{chang2023hydra}, and S-Graph+ \cite{bavle2023s} enabled real-time, multi-robot, and LiDAR-based construction of hierarchical 3D scene graphs, respectively. Recent works further incorporate open-vocabulary perception and foundation models to improve semantic generalization \cite{gu2024conceptgraphs,werby2024hierarchical,maggio2024clio}.
However, existing hierarchical 3D scene graphs mainly organize environments across scene-level scales. %, such as geometry, objects, places, rooms, and buildings. 
Objects are typically treated as atomic semantic nodes or represented by raw segments, meshes, point clouds, or simple primitives, with limited modeling of their multi-level structure. 
\arxivAdd{In contrast, our work introduces a hierarchical object-centric representation that preserves detailed geometry while providing compact superquadric abstractions for localization and navigation.}

\myParagraph{Global localization} 
Global localization estimates robot pose within a pre-built map, which is challenging in large-scale or repetitive environments. Visual place recognition methods, including NetVLAD \cite{arandjelovic2016netvlad}, Patch-NetVLAD \cite{hausler2021patch}, and DINOv2 \cite{oquab2023dinov2}, retrieve candidate locations using global descriptors. However, they mainly rely on appearance similarity and lack explicit geometric reasoning, making them vulnerable to perceptual aliasing.
Scene graph-based methods introduce semantic nodes and spatial relationships to improve structural reasoning. Hydra \cite{hughes2022hydra}, Hydra-Multi \cite{chang2023hydra}, and semantic graph-based place recognition \cite{kong2020semantic} improve robustness but require accurate perception and add system complexity.
Object-based localization uses objects as landmarks and aligns maps through object-level data association. Classical RANSAC-based methods suffer from combinatorial complexity and outlier sensitivity \cite{matsuzaki2024clip,matsuzaki2024clip2}. Recent methods such as CLIPPER \cite{lusk2021clipper} and ROMAN \cite{peterson2025roman} improve robustness through consistency maximization, but their performance remains limited by the expressiveness of the underlying object representation.

\textbf{Safe Navigation.} 
Safe navigation requires efficient collision avoidance and fast distance evaluation between robots and obstacles. Classical methods use iterative primitive-based algorithms such as GJK \cite{gilbert1988gjk} or sampling-based planners with repeated collision queries \cite{lavalle2006planning}, which can become expensive in cluttered scenes.
Dense scene representations, including point clouds, occupancy grids, and meshes, usually perform collision checking through nearest-neighbor searches or SDF queries \cite{oleynikova2017voxblox,newcombe2011kinectfusion}. Although accurate, their cost scales with scene complexity. Clearance constraints are often enforced by obstacle inflation or Minkowski sums \cite{lozanoperez1983spatial}, but these can increase computation and lead to conservative behavior.
Recent implicit and analytical representations, such as SDFs and parametric shapes \cite{park2019deepsdf,paschalidou2019superquadrics}, enable efficient inside-outside evaluation without iterative distance queries. However, they are often limited to simple geometries or are not fully integrated into object-level SLAM frameworks.

\section{Hierarchical Object Representation}
We now describe the hierarchical object representation in detail. See Figure~\ref{overview}.

\myParagraph{Layer 1: Point Cloud}
% \subsection{Layer 1: Point Cloud}
The first layer stores an object-level point cloud for each object instance. This layer provides the most direct geometric evidence from RGB-D observations and preserves the raw 3D support of the segmented object. It serves as the basic spatial observation layer for object localization, coarse geometric reasoning and temporal object registration. Compared with monolithic scene-level point clouds, the object-wise organization makes the representation more structured and directly assiciates geometry with semantic object identities.

\myParagraph{Layer 2: 3D Mesh}
% \subsection{Layer 2: 3D Mesh}
The second layer represents each object as a high-fidelity mesh. Compared with point clouds, meshes provide continuous surfaces and geometric and texture details, make them suitable for visualization and object-level scene reconstruction. This layer preserves rich shape information that may be lost in compact abstractions. However, mesh representations are memory-intensive and less efficient for vertices calculation, therefore motivating the higher-level analytical layers.

\myParagraph{Layer 3: Superquadric}
% \subsection{Layer 3: Superquadric}
The third layer abstracts each object mesh into a set of superquadric primitives. Each primitive compactly encodes an object's local geometry using a small number of scale, shape and pose parameters. This layer significantly reduces storage while retaining the dominant object geometry. More importantly, the analytical form of superquadrics enables efficient computation of object centroid and volume computation. It also allows us to efficiently extract pose-invariant shape descriptors. Therefore, this layer supports object association and multi-session map alignment through compact geometric comparison without relying on dense point cloud or mesh-level registration.

\myParagraph{Layer 4: Inflated Superquadric}
% \subsection{Layer 4: Inflated Superquadric}
The fourth layer further converts the fitted superquadrics in Layer 3 into inflated superquadrics. This layer is designed for conservative collision reasoning. By expanding the analytical primitives to cover the object geometry with an optional safety margin, the representation provides a compact approximation of the occupied configuration space. During navigation, collision checking can be reduced to evaluating the inside-outside function of the inflated primitives, avoiding expensive iterative distance queries against dense meshes or point clouds. This makes the inflated superquadric layer suitable for efficient robot planning and safe navigation in cluttered scenes.

\section{Building the Hierarchical Object Representation}
% \section{Methodology}
% \subsection{System Overview}
% Our proposed framework is designed for object-level scene reconstruction and representation for robust map alignment \& robot navigation. As illustrated, the system takes an RGB-D sequence of a novel scene and its corresponding camera trajectory as input to build a semantically rich and object-centric 3D map.

% The pipeline consists of three primary stages. First, the input stream is processed to extract and associate open-set 2D object masks across frames, establishing temporally consistent object identities. Second, to improve reconstruction accuracy and ensure computational efficiency, we propose a best-view strategy for each object to generate a high-fidelity 3D mesh and pose estimation in the camera frame. Finally, these dense representations are parameterized into superquadrics. This compact mathematical formulation serves a dual downstream purpose: first, it enables highly efficient shape comparison for robust data associations and odometry map alignment across sequences; second, it provides analytical closed-form evaluation of collision boundaries, directly outputting conservative configuration for safe navigation.

\begin{figure}[t]
\centering
\includegraphics[width=\linewidth]{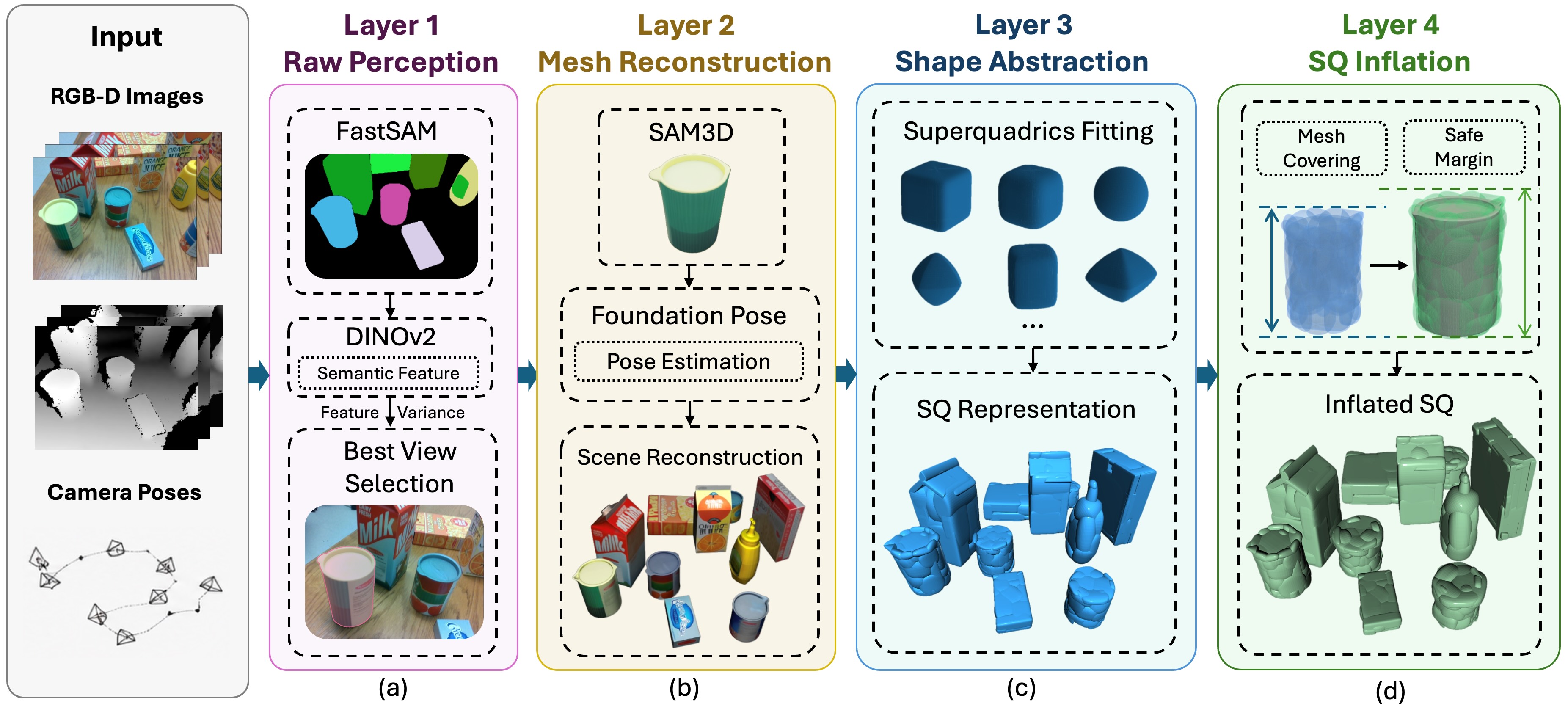}
\arxivRemove{\vspace{-7mm}}
\arxivAdd{\vspace{-4mm}}
\caption{\hickory: Pipeline for building the hierarchical object representation from posed RGB-D image stream. (a) Layer 1 performs object segmentation and feature-variance-based best-view selection. (b) Layer 2 reconstructs object meshes and estimates their poses to build an object-level scene. (c) Layer 3 abstracts each object mesh into a compact superquadric representations and shows its applicability in map alignment. (d) Layer 4 inflates the fitted superquadrics to conservatively cover object geometry for efficient collision checking and robot navigation.}
\label{pipeline}
\arxivRemove{\vspace{-5mm}}
\arxivAdd{\vspace{-4mm}}
\end{figure}

We develop a pipeline to build the hierarchical object representation from posed RGB-D image stream. See Figure~\ref{pipeline} for an overview. We describe the pipeline as it builds the different layers of the hierarchical representation.

% The framework builds a hierarchy of representations from dense observations to compact analytical abstractions. 
% Unlike directly relying on sparse pointclouds, it progressively converts RGB-D observations into object-level meshes, superquadric abstractions, and inflated SQs. 
% This hierarchy preserves geometric fidelity for scene reconstruction while enabling efficient downstream reasoning, including object association, map alignment, and collision-aware robot navigation.

\myParagraph{Layer 1: Open-Set Perception and Best-View Selection}
%\subsection{Layer 1: Open-Set Perception \& Best-View Selection}
%
%Our open-set perception module follows the mapping framework of ROMAN~\cite{peterson2025roman} while introducing modifications for object-level reconstruction. 
%Given an RGB-D sequence, we first use FastSAM~\cite{zhao2023fast} to generate class-agnostic instance masks for each RGB frame. To avoid reconstructing uninformative or unstable regions, OneFormer~\cite{jain2023oneformer} is used to reject predefined structural (\eg walls, floors, ceilings, sky, buildings) and dynamic classes (\eg persons). For each mask, DINOv2~\cite{oquab2024dinov2} features are extracted as pixel-level semantic descriptors for later view selection and object association.
%
%Each valid 2D mask is lifted into 3D using the registered depth map and camera intrinsics. After filtering unreliable depth points, the local point cloud is transformed into the global frame using the known camera pose $T_W^C \in SE(3)$. We then associate observations over time using incremental oriented bounding boxes (OBBs): each new observation is matched against existing object tracks by 3D IoU, updating the best-matched object if the IoU exceeds a threshold, or initializing a new object otherwise. This produces a temporally consistent object library with multi-view observations for each reconstructed instance. 
%
We generate class-agnostic instance object masks for each RGB keyframe using FastSAM~\cite{zhao2023fast} and extract pixel-level DINOv2~\cite{oquab2024dinov2} features. To avoid reconstructing uninformative or unstable regions, we reject predefined structural (\eg walls, floors, ceilings, sky, buildings) and dynamic classes (\eg persons) using a pre-trained OneFormer~\cite{jain2023oneformer}. 
Each valid 2D mask is lifted into 3D using the registered depth map and camera intrinsics. After filtering unreliable depth points, the local point cloud is transformed into the global frame using the known camera pose $\MT_W^C \in \SEthree$. We track objects over time using incremental oriented bounding boxes (OBBs): each new object segment is matched against existing object tracks by 3D intersection-over-union (IoU), updating the best-matched object if the IoU exceeds a threshold, or initializing a new object otherwise. This produces a temporally consistent object library with multi-view observations for each reconstructed object instance.

Running 3D object reconstruction models (\eg SAM3D) on every keyframe is computationally intensive and takes too much time. %Therefore, 
%rather than reconstructing the object continuously, 
Our pipeline selects a single \emph{best view} with the highest S-score for each registered object instance to perform high-fidelity mesh reconstruction. The S-score is given by
\begin{equation}
S = \ln(1 + \sigma^2) \cdot (R_{\mathit{area}})^\alpha \cdot (R_{\mathit{depth}})^\beta,
\end{equation}
where $\sigma$ denotes the variance in pixel-level DINOv2 features, $R_{\mathit{area}} = \frac{N}{H \times W}$ measures the visible mask area relative to the image size, and $R_{\mathit{depth}}$ denotes the ratio of valid depth pixels within the mask after depth filtering. 
\arxivAdd{Here, $R_{\mathit{area}}$ captures image-space object visibility, while $R_{\mathit{depth}}$ measures depth reliability; thus, a large mask with many invalid depth pixels can still be penalized.}
The exponents $\alpha$ and $\beta$ are empirical weighting coefficients controlling the contribution of visibility and depth reliability.
The key intuition is that an informative view should expose diverse and discriminative object parts, rather than only a flat, occluded, or textureless region. Since DINOv2 features encode local semantic and appearance variations, a view with higher feature variance $\sigma$ will contain richer object structures and more complete visible geometry.

\myParagraph{Layer 2: Mesh Reconstruction and Pose Estimation}
%\subsection{Layer 2: Mesh Reconstruction and Pose Estimation}
%
For each tracked object, our pipeline generates a complete textured 3D mesh from the selected best view using SAM3D~\cite{chen2025sam}. 
The estimated poses by SAM3D tend to be not very accurate. 
Our pipeline uses reconstructed mesh as the geometric template for FoundationPose~\cite{wen2024foundationpose} to estimate more reliable object pose in the local camera frame. 
Finally, using the known camera trajectory, the estimated object pose is transformed into the world frame. 

%We use the selected best view for each object to gene
%Given the selected best view of an object, we input the corresponding segmented mask and depth image to SAM3D~\cite{chen2025sam} to generate a complete metric-scale mesh. Compared with point cloud representations, the reconstructed mesh provides a continuous surface representation that preserves fine geometric details while remaining compact for storage and visualization.
%
%We then use the reconstructed mesh as the geometric template for FoundationPose~\cite{wen2024foundationpose} to estimate the object pose in the local camera frame. By leveraging both the generated geometry and RGB-D observations, FoundationPose produces a robust object pose estimate even under partial observations and viewpoint changes.
%
%Finally, using the known camera trajectory, the estimated object pose is transformed into the world frame. This allows all reconstructed objects to be assembled into a consistent object-level scene representation. As illustrated in Fig.~\ref{pipeline}(b), the resulting scene preserves both metric-scale geometry and object poses, providing the foundation for the compact shape abstraction introduced in the next layer. \RT{Can be omited.}

\myParagraph{Layer 3: Shape Abstraction and Superquadric Fitting}
%\subsection{Layer 3: Superquadric Representation}
%
%Traditional object-centric reconstruction often represent objects using either partial point clouds, which are memory-intensive and sensitive to occlusions, or simple 3D bounding boxes, which are fundamentally too coarse to capture meaningful geometric details. To seamlessly integrate the fine-grained reconstruction output from SAM3D into downstream applications, it is imperative to abstract the dense 3D meshes into a compact and mathematically expressive parameterization. To this end, we utilize superquadrics as our geometric primitives.
%
% \begin{figure}[t]
% \centering
% \includegraphics[width=0.9\linewidth]{figures/fig_3.png}
% \caption{Example of Superquadrics and Shape Abstract by SQ Fitting \cite{liu2023marching}.}
% \label{fig3}
% \end{figure}
%
A superquadric is a parameterized shape model and its shape surface can be described using an analytical expression: 
\begin{equation}
    G(\vp) = \left( \left| \frac{x}{a_1} \right|^{\frac{2}{\epsilon_2}} + \left| \frac{y}{a_2} \right|^{\frac{2}{\epsilon_2}} \right)^{\frac{\epsilon_2}{\epsilon_1}} + \left| \frac{z}{a_3} \right|^{\frac{2}{\epsilon_1}} = 1,
    \label{eq:sq}
\end{equation}
where $\vp =(x, y, z)$ denotes a 3D point, $a_1$, $a_2$ and $a_3$ govern the spatial scale of the primitive along the three principal axes, while $\epsilon_2$ controls the shape in the $xy$ plane and $\epsilon_1$ controls the shape along the vertical direction. Recent works have successfully investigated ways to fit a 3D object shape with a series of superquadric primitives \cite{liu2023marching}, thereby providing memory-efficient representation of the accurate 3D object geometry. 

Our pipeline uses Marching-Primitives \cite{liu2023marching} to decompose the dense 3D mesh in Layer 2 into a set of superquadric primitives. We observe that the variability in shape estimation across different views and robot trajectories do not significantly affect the first set of superquadric primitives. This allows us to use the parameters of the first $K$ fitted superquadrics as an efficient, invariant shape representation of the object. The Layer 3 stores the first $K$ superquadric primitives for each object. During object comparison, we define $K$ according to the number of superquadrics of the object with fewer fitted primitives.

%While a single superquadric provides a high memory-efficient descriptor, it fundamentally lacks the degrees of freedom required to accurately capture the intricate geometric details of complex objects. To address this, we adopt the shape abstraction approach introduced by Marching-Primitives \cite{liu2023marching}. This method decomposes a dense mesh into different components and fit them with a set of superquadric primitives, then iteratively optimizes the parameters of superquadrics to minimize its geometric discrepancy with the target local shape. By repeating the process for each object, we can build an light scene representation shown in Fig. \ref{pipeline}(c) that largely reduces the memory sizes while retaining sufficient geometric expressiveness.

\emph{Data Association \& Map Alignment.} We utilize the Layer 3 (\ie first $K$ superquadric primitive parameters) for object-based map alignment. 
In it, we are given two maps $\mathcal{M}_A$ and $\mathcal{M}_B$ with Layer 1-3 representation of the objects. 
The problem is to identify the same object instances in the two maps, and align the map by obtaining the relative pose $\MT_A^B$ between the two maps.
We follow the same approach as in ROMAN \cite{peterson2025roman}, which finds data association by assigning unary scores to each putative object-object association and binary scores for each pair of putative object-object associations. ROMAN then uses a prior work CLIPPER \cite{lusk2021clipper} to identify the inlier object-object associations, and obtains map alignment by solving a simple registration problem. 

ROMAN proposes semantic, geometric, and gravity scores for a putative object-object association. We implement the same approach, however, only modify the geometric score $s_{\mathit{geo}}$. ROMAN uses a geometric score that is extracted out of the point cloud (Layer 1). We instead propose the following geometric score $s_{\mathit{geo}}$ based on superquadric parameterization in Layer 3:
\begin{equation}
    s_{\mathit{geo}}
    =
    \sum_{k=1}^{K}
    w_k
    \cdot
    \exp\left(
    -\frac{
    |\epsilon_{1}^{k}-\bar{\epsilon}_{1}^{k}|
    +
    |\epsilon_{2}^{k}-\bar{\epsilon}_{2}^{k}|
    }{\sigma_{\mathit{geo}}}
    \right)
    \cdot
    \left(
    \frac{1}{3}
    \sum_{j=1}^{3}
    \frac{
    \min(a_j^k,\bar{a}_j^k)
    }{
    \max(a_j^k,\bar{a}_j^k)
    }
    \right),
    \label{sgeo}
\end{equation}
where
$\{(\epsilon_1^k,\epsilon_2^k,a_1^k,a_2^k,a_3^k)\}_{k=1}^{K}$
and
$\{(\bar{\epsilon}_1^k,\bar{\epsilon}_2^k,\bar{a}_1^k,\bar{a}_2^k,\bar{a}_3^k)\}_{k=1}^{K}$
denote the Layer 3 superquadric parameterizations of two objects in maps $\mathcal{M}_A$ and $\mathcal{M}_B$, respectively.
Here, the (second) exponent term (involving $\epsilon^k_{i}$, $\bar{\epsilon}^k_{i}$) penalizes shape discrepancy, where as the (third) average term penalizes scale alignment across the three principal axis. The weights $w_k$ ensures volume-weighted average in~\eqref{sgeo}, and equals the ratio of the volume of superquadric $k$ and the total volume of the object. We compute volumes using analytical formulas available for superquadrics~\cite{jaklic2000segmentation}.

\myParagraph{Layer 4: Inflated Superquadrics and Collision Checking for Robot Navigation}
%\subsection{Layer 4: Superquadrics for Collision Avoidance \& Robot Navigation}
%
To construct a conservative configuration space for robot navigation, we first optionally inflate the object point cloud (Layer 1) and object mesh vertices (Layer 2) outwards from the object centroid by a predefined safety margin, $r_{safe}$. We use the centroid computed using an analytical formula using superquadric fit (Layer 3)~\cite{jaklic2000segmentation}. 
To inflate the superquadrics, 
we assign each inflated point and mesh vertex to its geometrically closest superquadric primitive. 
Let $\calP_k$ denote the set of inflated points and mesh vertices closest to the superquadric primitive $k$ and $G_k(\cdot)$ denote the surface function in~\eqref{eq:sq} for the primitive $k$. 
Then, for each superquadric primitive $k$, we evaluate the function $G_k(\vp)$ for all $\vp \in \calP_k$ and compute $G_{\max,k}=\max_{\vp \in \calP_k}G_k(\vp)$. We scale the superquadric along each of its principal axes so that $G_k(\vp) \leq g$ for all $\vp \in \calP_k$, where $g \leq 1$ is a target IOF value that controls the conservativeness of the inflated primitive.
The geometric scale factor that ensures this condition is given by
\begin{equation}
\gamma_k = \left(\frac{G_{\max,k}}{g}\right)^{\frac{\epsilon_{1,k}}{2}}.
\end{equation}
Repeating this process for all primitives a yields a hierarchical representation shown in Fig. \ref{pipeline}(d) that fully encompasses the physical extents of the object augmented by a safety margin.

\emph{Analytical Collision Checking for Robot Navigation.} 
During real-time path planning, 
%the robot then can be regarded as a point mass expanded with an extra radius margin. 
we do collision checking analytically using the inflated superquadrics (Layer 4).
For any queried 3D point $\vp$, we transform it into the local frame of the superquadric primitives. The spatial point is mathematically guaranteed to be collision-free if it lies strictly outside all the inflated superquadrics, which reduces to a purely algebraic evaluation:
\arxivAdd{\begin{equation}
     G_{k}(\MT_k^{-1} \tilde{\vp})  > 1~~\text{for all}~~k,
\end{equation}}
\arxivRemove{$G_{k}(\MT_k^{-1} \tilde{\vp})  > 1~~\text{for all}~~k,$}
where $\tilde{\vp} \triangleq [\vp\tran 1]\tran$.
This reduces the complex collision checking problem to a light and fast scalar check yielding benefits in dense and cluttered environments.

\section{Experiment}

We comprehensively evaluate the proposed pipeline across three key dimensions. We first assess the geometric fidelity of our object-level scene reconstruction by our best-view selection against a series of strategies (Sec.\ ~\ref{sec:exp_reconstruction}). Second, we validate the robustness of our object representation in local map alignment across different sequences (Sec.\ ~\ref{sec:exp_alignment}). Third, we evaluate the path planning using our analytical representation to showcase its effectiveness in robot navigation (Sec.\ ~\ref{sec:exp_navigation}). 
%We conduct experiments on synthetic and real-world datasets such as ReplicaCAD, HOPE,  Kimera-Multi, and our own NUS Campus (NUS-CLB) collected on-board Unitree B2 quadruped robot.
To validate the applicability and robustness of our pipeline in real-world scenarios, we deploy the pipeline for downstream tasks on a Unitree B2 quadruped robot equipped with a ZED 2i RGB-D camera.
For all experiments, we use a workstation with Intel i9-14900KF CPU and NVIDIA Geforce RTX 4090 GPU.

\subsection{Hierarchical Object Representation: Memory and Time}
We first report the processing time for building each layer in our representation and the average storage required by each layer. 
See Table \ref{tab:representation_storage_runtime}.
We observe that dense 3D mesh (Layer 2) consumes the most storage and has the highest processing time. This is because Layer 2 stores fine-grained textured mesh. The high run-times are caused primarily SAM3D. 
We observe that the memory and processing time requirements for Layer 3 and 4 remain very manageable for real-world deployment.
\begin{table}[htbp]
\arxivRemove{\vspace{-0.4cm}}
\centering
\caption{Average storage and processing time for building each layer of our hierarchical object representation across datasets.
$\bar{N}$ is the average number of objects reconstructed.}
\vspace{-2mm}
\label{tab:representation_storage_runtime}
\small
\renewcommand{\arraystretch}{1.15}
\begin{tabular}{lcccc}
\toprule
\textbf{Layers} &
\makecell[c]{\textbf{HOPE}\\$(\bar{N}=16.67)$} &
\makecell[c]{\textbf{ReplicaCAD}\\$(\bar{N}=42.25)$} &
\makecell[c]{\textbf{Kimera-Multi}\\$(\bar{N}=41.83)$} &
\makecell[c]{\textbf{NUS-CLB}\\$(\bar{N}=40)$} \\
\midrule
Layer 1: Point cloud   
& \makecell[c]{61.04 KiB\\(61.32 s)}
& \makecell[c]{408.61 KiB\\(124.61 s)}
& \makecell[c]{703.26 KiB\\(231.34 s)}
& \makecell[c]{471.69 KiB\\(182.83 s)} \\

Layer 2: 3D Mesh       
& \makecell[c]{232.46 MiB\\(167.63 s)}
& \makecell[c]{437.12 MiB\\(345.12 s)}
& \makecell[c]{324.69 MiB\\(349.28 s)}
& \makecell[c]{483.84 MiB\\(392.51 s)} \\

Layer 3: Superquadrics 
& \makecell[c]{51.12 KiB\\(68.12 s)}
& \makecell[c]{73.34 KiB\\(165.23 s)}
& \makecell[c]{62.78 KiB\\(151.67 s)}
& \makecell[c]{52.48 KiB\\(172.26 s)} \\

\makecell[l]{Layer 4: Inflated \\ Superquadrics} 
& \makecell[c]{51.12 KiB\\(17.12 s)}
& \makecell[c]{73.34 KiB\\(43.92 s)}
& \makecell[c]{62.78 KiB\\(38.26 s)}
& \makecell[c]{52.48 KiB\\(35.36 s)} \\

% \midrule
% \textbf{Total}
% & \makecell[c]{232.57 MiB\\(314.19 s)}
% & \makecell[c]{437.60 MiB\\(678.88 s)}
% & \makecell[c]{325.44 MiB\\(770.55 s)}
% & \makecell[c]{484.40 MiB\\(782.96 s)} \\
\bottomrule
\end{tabular}
\arxivRemove{\vspace{-0.9cm}}
\arxivAdd{\vspace{-4mm}}
\end{table}

\subsection{Evaluation on Object \& Scene Reconstruction}
\label{sec:exp_reconstruction}

\emph{Datasets and Baselines.} To evaluate the reconstruction fidelity of our proposed pipeline, we conduct reconstruction experiments using HOPE \cite{tyree2022hope} and ReplicaCAD \cite{szot2021habitat} datasets. Specifically, for HOPE dataset, we evaluate our pipeline across 3 distinct scenes, each containing more than 10 diverse objects. For ReplicaCAD, we run the simulation to record camera trajectories across 4 indoor apartment scenes, and both datasets provide the groundtruth object meshes and their 6D poses. To validate the effectiveness of our proposed best-view selection, we compare it against four heuristic baselines: \textit{Initial-View}, which selects the frame where the object is first registered; \textit{Max-Area (Mask)}, which uses the frame with the highest 2D segmentation pixel count; \textit{Max-BBox (2D)}, which chooses the largest 2D bounding box area; and \textit{Max-Volume (3D)}, which selects the frame yielding the maximum 3D bounding box volume from the projected point cloud. For each object reconstructed by SAM3D with different selections, we evaluate the accuracy against the groundtruth. The geometric fidelity is measured using F1 score and the pose estimation is evaluated by the symmetric Average Distance (ADD-S) metric.

\arxivAdd{\begin{table}[b] 
\vspace{-5mm}}
\arxivRemove{\begin{table}}
\centering
\caption{Quantitative evaluation of object shape and pose. We report the average accuracy across all scenes for each dataset. \textbf{Bold} indicates the best performance, while \underline{underlined} indicates the second best. $\uparrow$ denotes higher is better.}
\vspace{-2mm}
\label{tab:reconstruction_avg}
\begin{tabular}{l cccc}
\toprule
\multirow{2}{*}{\textbf{Method}} & \multicolumn{2}{c}{\textbf{HOPE (Avg.)}} & \multicolumn{2}{c}{\textbf{ReplicaCAD (Avg.)}} \\
\cmidrule(lr){2-3} \cmidrule(lr){4-5}
& \textbf{Mean $F_1$} $\uparrow$ & \textbf{ADD-S} $\uparrow$ & \textbf{Mean $F_1$} $\uparrow$ & \textbf{ADD-S} $\uparrow$ \\ 
\midrule
Initial-View & 0.9141 & 0.7600 & 0.8395 & 0.7317 \\
Max-Area (Mask) & \underline{0.9215} & 0.8200 & 0.8566 & \underline{0.7886} \\
Max-BBox (2D) & 0.8977 & \underline{0.8300} & 0.8619 & 0.7628 \\
Max-Volume (3D) & 0.8472 & 0.7000 & \underline{0.8649} & 0.7398 \\
\midrule
\textbf{Ours (View-Score)} & \textbf{0.9312} & \textbf{0.8400} & \textbf{0.8721} & \textbf{0.7967} \\
\bottomrule
\multicolumn{5}{l}{\footnotesize Note: ADD-S thresholds are @0.05d (HOPE) and @0.1d (ReplicaCAD).} \\
\end{tabular}
\arxivRemove{\vspace{-6mm}}
\end{table}

\emph{Results and Discussion.} As demonstrated in the Table \ref{tab:reconstruction_avg}, our proposed View-Score selection consistently achieves the highest performance in both geometric fidelity (Mean $F_1$) and pose estimation accuracy (ADD-S) across the two datasets. Baselines relying solely on 2D image footprints, such as Max-Area (Mask) and Max-BBox (2D), inherently degenerate into selecting the camera view physically closest to the object. While this maximizes the pixel count, it frequently captures a frontal view that severely lacks critical depth and side-profile information. Conversely, the Max-Volume (3D) baseline attempts to capture the struggles with thin or planar objects, for which it often misses essential geometric parts due to object's uneven distribution along all principal axes and struggles with perception noises. 
By utilizing feature variance, and balancing it with the mask area, our score actively prioritizes frames with a higher diversity of structural details. 
%Consequently, the selected optimal view provides a structurally complete partial point cloud, ensuring that the subsequently reconstructed mesh accurately encapsulates the full topology and also benefiting the pose estimation.

\subsection{Evaluation on Object Association \& Map Alignment}
\label{sec:exp_alignment}

\emph{Datasets and Baselines.} We now validate the 
%The second experiment is to validate 
our Superquadric-based association score for robust map alignment. As the datasets visualized in Fig. \ref{align_data}, we perform cross-sequence map alignment on ReplicaCAD \cite{szot2021habitat} (simulated indoor) and Kimera-Multi  \cite{tian2022kimera} (real-world outdoor) datasets. For each of the four indoor scenes in ReplicaCAD, we use the simulator to record two distinct RGB-D camera trajectories. For the large-scale Kimera-Multi dataset, we leverage three pairs of odometry sequences captured by two mobile robots that exhibit partial trajectory overlap for association. %Upon reconstructing the local map for each sequence, we compute the relative transformation between the two odometry frames by the association method. This estimated transformation is evaluated against the true value derived from the groundtruth of global poses of the two sequences. 
\begin{figure}[t!]
\centering
\includegraphics[width=1.0\linewidth]{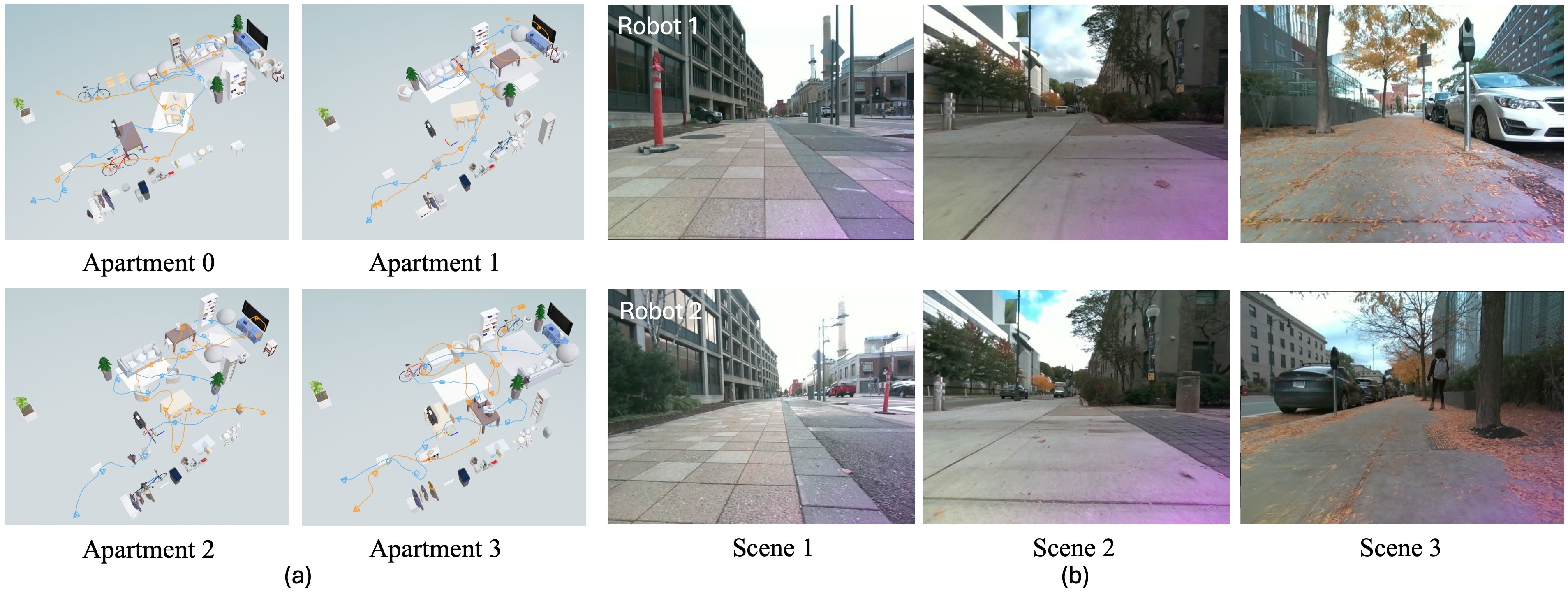}
\arxivRemove{\vspace{-6mm}}
\arxivAdd{\vspace{-4mm}}
\caption{Visualization of Dataset for Map Alignment Evaluation: (a) Four indoor scenes in ReplicaCAD \cite{szot2021habitat}, featuring distinct object configurations. The two independent camera trajectories for each scene are highlighted in blue and orange. (b) Three outdoor scenes in Kimera-Multi \cite{tian2022kimera}, where two robots traverse shared locations either in the same direction (Scenes 1 and 2) or in opposite directions (Scene 3). Note that dynamic objects may vary across sequences due to temporal differences between recordings.}
\label{align_data}
\arxivRemove{\vspace{-0.5cm}}
\arxivAdd{\vspace{-4mm}}
\end{figure}
We compare our approach with ROMAN \cite{peterson2025roman}, a state-of-art baseline for object-based map alignment. %that relies on partial point clouds and OBB features for shape matching. 
We chose the version \textit{ROMAN-XL} with the best reported performance.
ROMAN-XL divides the scene into submaps and align them by submap-pairs. 
We do not use submap construction. In order to make the coparisons fair, we also compare against \textit{ROMAN-One}, that builds the whole scene in a single map without submap division. Furthermore, we introduce a \textit{Mesh-BBox} baseline, which performs association using the bounding box of the complete meshes generated by SAM3D. %Besides the shape score, all method include the semantic feature and the spatial matching terms for overall score calculation.

\emph{Results and Discussion.} 
\arxivAdd{An example of the map alignment in the confined ReplicaCAD environment is and NUS Campus Dataset (NUS-CLB) is shown in Fig. \ref{align_vis}.}
As detailed in Table \ref{tab:reloc_eval_full}, our superquadric-based matching achieves accurate map alignment across the tested datasets. Conversely, the ablation baseline, Mesh-BBox, exhibits significantly higher errors. Because its association relies on the coarse dimensions of mesh bounding boxes, it fundamentally fails to distinguish geometrically different objects that coincidentally share similar spatial volumes, leading to brutally wrong matching and degrading its performance especially under scenes with fewer objects. 

\arxivAdd{
\begin{figure}[t]
\centering
\includegraphics[width=0.95\linewidth]{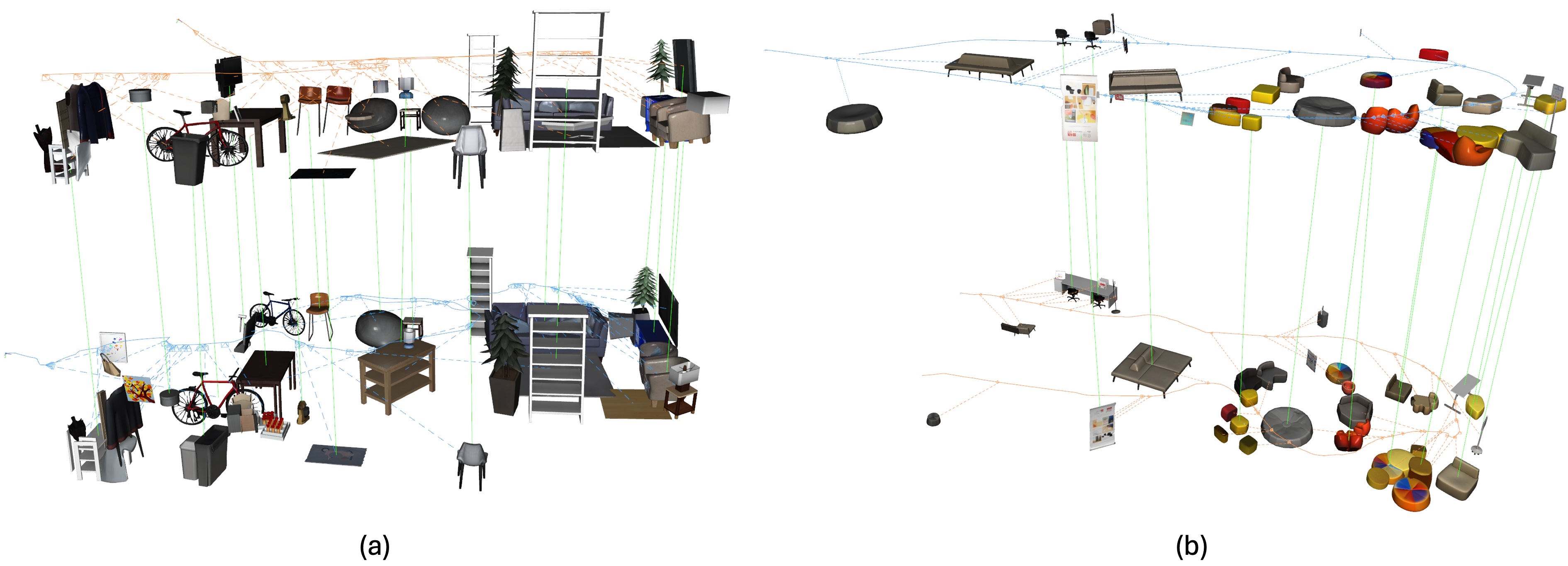}
\vspace{-2mm}
\caption{Visualization of Map alignment Result. (a) Evaluation in simulation on ReplicaCAD dataset. (b) Demonstration in real world on NUS-CLB scene.}
\vspace{-4mm}
\label{align_vis}
\end{figure}
}

When compared against the ROMAN series, our results reveal a critical insight regarding the \textit{quality versus quantity} of object associations. While constructing a monolithic global map (ROMAN-One) theoretically increases the raw number of matches ($N_a$) and yields competitive indoor accuracy, this strategy proves highly brittle in real-world environments. In the large-scale, sensor-noisy Kimera-Multi dataset, ROMAN-One's blind pursuit of associations leads to a great number of false-positive matching, as evidenced by its $3.26$ m translation error in Sce 3. In contrast, our method achieves state-of-the-art accuracy while maintaining a concise association count. This confirms that the intrinsic quality of our superquadric-based associations is significantly higher. We attribute this to the fact that point cloud methods like ROMAN-XL frequently mismatch fragmented object segments. Our pipeline, however, leverages the best-view selection for reconstruction and hierarchical superquadric representation to help strutural completeness, ensuring that associations are grounded in the full geometric association of the objects rather than ambiguous partial surface overlaps.

For real-world deployment, we evaluate the system in a large and cluttered library environment. We report examples of map alignment in both real-world dataset and deployment scenarios in the supplementary video. 

\begin{table}
\centering
\caption{Quantitative evaluation of map alignment across different datasets. We report detailed matching numbers ($N_a$) as an objective reference, along with translation and rotation errors. \textbf{Bold} indicates the best performance, and \underline{underlined} indicates the second best for error metrics. $\downarrow$ denotes lower is better.}
\label{tab:reloc_eval_full}
\resizebox{\textwidth}{!}{
\begin{tabular}{ll ccccc cccc}
\toprule
\multirow{2}{*}{\textbf{Method}} & \multirow{2}{*}{\textbf{Metric}} & \multicolumn{5}{c}{\begin{tabular}[c]{@{}c@{}}\textbf{ReplicaCAD} \\ \textbf{(Indoor / Simulated)}\end{tabular}} & \multicolumn{4}{c}{\begin{tabular}[c]{@{}c@{}}\textbf{Kimera-Multi} \\ \textbf{(Outdoor / Real-world)}\end{tabular}} \\
\cmidrule(lr){3-7} \cmidrule(lr){8-11}
& & \textbf{Apt 0} & \textbf{Apt 1} & \textbf{Apt 2} & \textbf{Apt 3} & \textbf{Avg.} & \textbf{Sce 1} & \textbf{Sce 2} & \textbf{Sce 3} & \textbf{Avg.} \\ 
\midrule

\multirow{3}{*}{Mesh-BBox} 
& Assoc. ($N_a$) & 21 & 28 & 24 & 24 & 24.25 & 15 & 24 & 10 & 16.33 \\
& Trans. (m) $\downarrow$    & 0.0503 & 0.1363 & 0.0363 & 0.0511 & 0.0685 & 0.4700 & \textbf{0.1270} & 1.6137 & \underline{0.7369} \\
& Rot. ($^\circ$) $\downarrow$ & 1.0367 & 2.0622 & \underline{0.4211} & \underline{0.3429} & 0.9657 & 2.6211 & 0.6898 & 8.3873 & 3.8994 \\ 
\midrule

\multirow{3}{*}{ROMAN-XL} 
& Assoc. ($N_a$) & 21 & 21 & 14 & 25 & 20.25 & 26 & 28 & 28 & 27.33 \\
& Trans. (m) $\downarrow$    & 0.0474 & \underline{0.0286} & 0.0646 & \textbf{0.0195} & 0.0400 & 0.4963 & 0.4558 & \underline{1.4948} & 0.8156 \\
& Rot. ($^\circ$) $\downarrow$ & 1.0280 & 1.3962 & 2.8784 & 1.5031 & 1.7014 & 0.2634 & \textbf{0.0806} & \textbf{0.2006} & \textbf{0.1816} \\ 
\midrule

\multirow{3}{*}{ROMAN-One} 
& Assoc. ($N_a$) & 47 & 38 & 50 & 43 & 44.50 & 58 & 36 & 27 & 40.33 \\
& Trans. (m) $\downarrow$    & \textbf{0.0301} & \textbf{0.0189} & \underline{0.0285} & 0.0438 & \textbf{0.0303} & \textbf{0.0789} & 0.5276 & 3.2629 & 1.2898 \\
& Rot. ($^\circ$) $\downarrow$ & \underline{0.6063} & \textbf{0.7057} & 0.4921 & 0.3822 & \underline{0.5466} & \underline{0.1697} & \underline{0.0884} & \underline{0.8326} & \underline{0.3635} \\ 
\midrule

\multirow{3}{*}{\textbf{Ours}}
& Assoc. ($N_a$) & 19 & 26 & 21 & 23 & 22.25 & 15 & 23 & 9 & 15.67 \\
& Trans. (m) $\downarrow$    & \underline{0.0421} & 0.0474 & \textbf{0.0244} & \underline{0.0437} & \underline{0.0394} & \underline{0.1117} & \underline{0.1275} & \textbf{0.4952} & \textbf{0.2448} \\
& Rot. ($^\circ$) $\downarrow$ & \textbf{0.3423} & \underline{0.8282} & \textbf{0.2325} & \textbf{0.1890} & \textbf{0.3980} & \textbf{0.1041} & 0.2580 & 1.3133 & 0.5585 \\ 
\bottomrule
\end{tabular}
}
\vspace{-2mm}
\end{table}

\subsection{Evaluation on Robot Navigation}
\label{sec:exp_navigation}
\emph{Setup and Baselines.} We now evaluate the downstream utility of our hierarchical representation in robot navigation in dense and cluttered environments.
We conduct these experiments in real-world setting using our Unitree B2 quadrupedal robot. 
We benchmark our method against two widely used baseline representations: a conservative \textit{Mesh-OBB} using object bounding boxes (OBB) of object meshes saved during scene reconstruction, and a \textit{Mesh-Full} baseline leveraging dense 3D mesh (Layer 2). 
To support efficient planning, the 3D scene is projected into a 2D occupancy grid by aggregating collision evaluations over heights within the robot's physical height. 
For Mesh-Full, object models must be loaded and explicitly rasterized onto the grid to represent obstacles. In contrast, our method elegantly determines free space directly evaluating the analytical implicit functions of the inflated superquadrics (Layer 4), bypassing the expensive rasterization entirely. We deploy a standard RRT planner to navigate between randomly sampled start and goal poses for 100 trials for each of 5 random seeds and compare the results for different representations. To normalize the immense variance inherent to random sampling, path quality is evaluated using the Mean Efficiency Ratio (straight-line Euclidean distance divided by planned path length).

% \begin{table}[htbp]
% \centering
% \caption{Quantitative evaluation of navigation performance. \textbf{Bold} and \underline{underlined} values indicate the best and second-best performance, respectively.}
% \label{tab:nav_eval}
% \resizebox{\columnwidth}{!}{
% \begin{tabular}{l ccccc}
% \toprule
% \textbf{Method} & \begin{tabular}[c]{@{}c@{}}\textbf{Succ. Rate}\\\textbf{(\%)} $\uparrow$\end{tabular} & \begin{tabular}[c]{@{}c@{}}\textbf{Eff. Ratio}\\ $\uparrow$\end{tabular} & \begin{tabular}[c]{@{}c@{}}\textbf{Map Build Time}\\\textbf{(s)} $\downarrow$\end{tabular} & \begin{tabular}[c]{@{}c@{}}\textbf{Map Storage}\\\textbf{(KiB)} $\downarrow$\end{tabular} & \begin{tabular}[c]{@{}c@{}}\textbf{Free Space}\\\textbf{Ratio} $\uparrow$\end{tabular} \\
% \midrule
% Mesh-OBB & 89.8 & 0.742 & \textbf{0.014} & \textbf{4.7} & 0.692 \\
% Mesh-Full & \textbf{100.0} & \textbf{0.893} & 209.192 & 1638.4 & \textbf{0.838} \\
% \midrule
% \textbf{Ours (SQ)} & \textbf{100.0} & \underline{0.885} & \underline{0.271} & \underline{77.9} & \underline{0.812} \\
% \bottomrule
% \end{tabular}
% }
% \end{table}

\begin{table}[t]
%\vspace{-6mm}
\centering
\caption{Quantitative evaluation of navigation performance. \textbf{Bold} and \underline{underlined} values indicate the best and second-best performance, respectively.}
\vspace{-2mm}
\label{tab:nav_eval}
\small
\setlength{\tabcolsep}{6pt}
\renewcommand{\arraystretch}{1.15}
\begin{tabular}{lcccc}
\toprule
\textbf{Method} &
\shortstack{\textbf{Success}\\\textbf{Rate (\%)} $\uparrow$} &
\shortstack{\textbf{Efficiency}\\\textbf{Ratio} $\uparrow$} &
\shortstack{\textbf{Occupy Map Build}\\\textbf{Time (s)} $\downarrow$} &
\shortstack{\textbf{Free Space}\\\textbf{Ratio} $\uparrow$} \\
\midrule
Mesh-OBB       & 89.8           & 0.742           & \textbf{0.014}  & 0.692 \\
Mesh-Full      & \textbf{100.0} & \textbf{0.893}  & 209.192         & \textbf{0.838} \\
\midrule
\textbf{Ours} & \textbf{100.0} & \underline{0.885} & \underline{0.271} & \underline{0.812} \\
\bottomrule
\end{tabular}
\vspace{-5mm}
\end{table}

\begin{figure}[t]
\centering
\includegraphics[width=\linewidth]{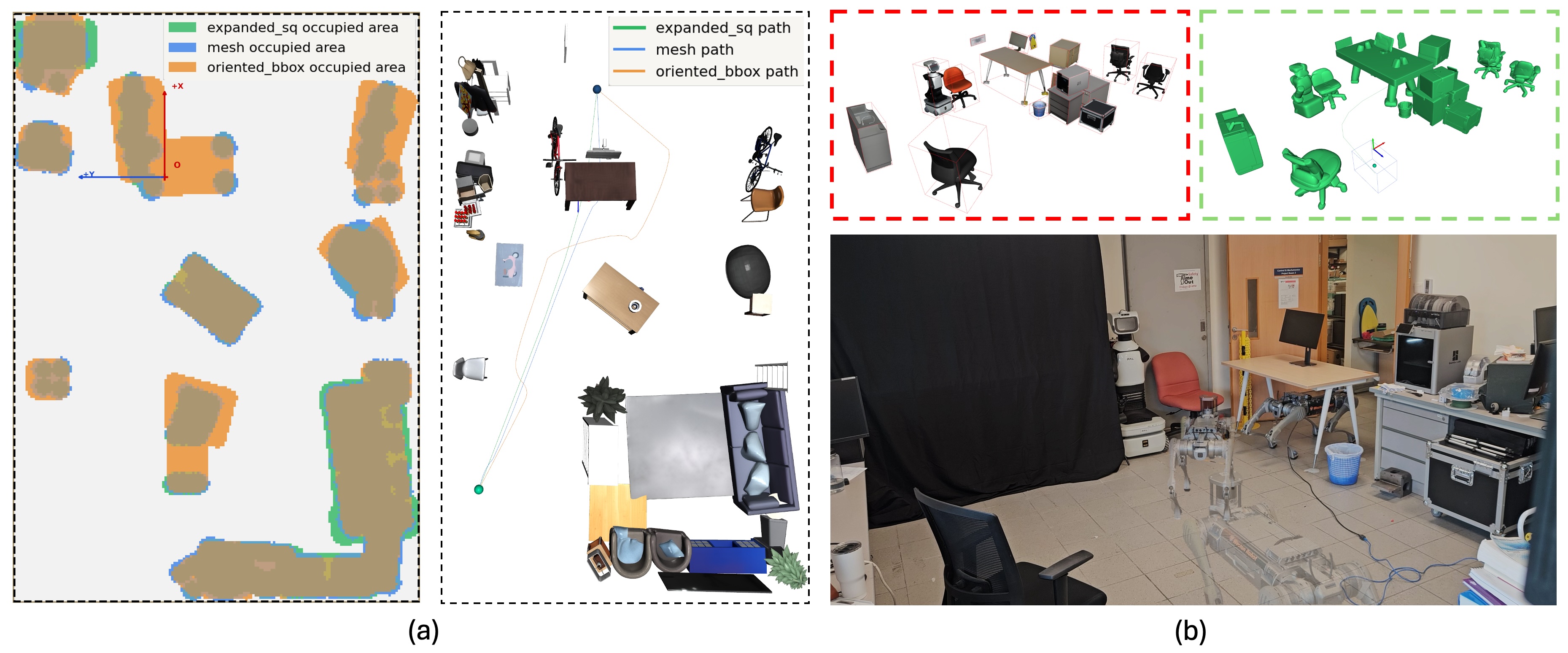}
\vspace{-7mm}
\caption{Evaluation on Robot Navigation. (a) Simulation evaluation. Comparison of 2D occupancy grid maps generated by different representation methods (left), and the corresponding planned navigation paths given the start and end positions (right). (b) Real-world demonstration. The occupy map by Mesh-OBB representation (red) fails to plan a valid path while our inflated superquadric representation (green) is able find path by \emph{ducking under the table to reach the goal} in real environment.}
\label{navi}
\vspace{-0.5cm}
\end{figure}

\emph{Results and Discussion.} The Mesh-Full baseline, while providing the highest free space ratio (0.838) and the most efficient paths (0.893), is time-consuming in generating the collision map via dense mesh rasterization and requires massive memory overhead. See Table \ref{tab:nav_eval}. Conversely, Mesh-OBB is computationally trivial, its severe over-approximation of object boundaries aggressively shrinks the navigable free space (0.692). This conservative padding causes the robot to fail to find valid paths like crossing under the table (dropping the success rate to 89.8\%) shown in Fig. \ref{navi}(a) and forces lengthy detours resulting in a degraded efficiency ratio of 0.742. 
Our method, which uses inflated superquadrics (Layer 4), achieves a balance between computational efficiency and geometric fidelity. 
The real-world robot execution is shown in Fig. \ref{navi}(b) and in the supplementary video. Our method is able to derive narrow paths in a confined indoor environment featuring a narrow passage (\eg navigating by ducking under the table). %By leveraging the inflated superquadric (Layer 4), the planner derives a collision-free configuration space. This explicitly enabled the robot to discover a safe trajectory through the narrow gap (e.g. through under the table).

\section{Conclusion}
%
% This paper proposes a hierarchical scene reconstruction framework from RGB-D sequences. For high-fidelity object reconstruction, we introduce best-view strategy to ensure geometrically rich and structurally complete mesh generation. To achieve a lightweight representation for downstream applications, we abstract these meshes into compact superquadrics. This enables robust multi-session map alignment via Superquadric-based object association and robot navigation using inflated superquadrics for analytical collision checking. Extensive evaluations validate the superiority of our framework over baseline methods, and demonstrations in real-world deployment illustrate its effectiveness and robustness against domain variations. For future work, to mitigate mask degradation caused by over-segmentation of FastSAM, we consider semantic-driven region merging to temporally fuse masks across frames. Besides, we aim to develop a scene-level parallelization in shape abstraction to overcome the runtime bottleneck of the current pipeline.
% RT -- Conclusion reads like summary + some future work that we did not do.
% RT -- Conclusion should read: what this work now enables
%
The paper proposes a hierarchical object representation that represents objects in four layers, from raw point clouds to dense 3D meshes to sparse parameterized superquadrics. We develop an automatic pipeline to build the hierarchical representation and show that it enables high-fidelity object-level reconstruction, robust object-level map alignment, and analytical collision checking for safe navigation planning. Our superquadric-based map alignment outperforms the current state-of-the-art ROMAN. Extensive evaluations on synthetic and real-world datasets and experiments on Unitree B2 Robot demonstrate the effectiveness of our pipeline and the usefulness of our representation.  

Objects are the first thing robots need to tackle when working in most real-world environments. 
%
% In this work, we argue that collating a mix of representations is helpful as different downstream task will require different 
% A hierarchical object representation helps 
This work considers the question of object representation in robot perception and posits a hierarchical object representation where each layer is designed to function for a specific downstream task. We believe that different downstream tasks will have different requirements, and such a mix of representation will help many more tasks in future works (\eg grasping and manipulation) and not just those considered here (\eg analytical collision checking for safe navigation, map alignment). Recent success in image-based object reconstruction models (\eg SAM3D) allows the research community to \arxivRemove{now investigate in this direction.}
\arxivAdd{now investigate in this direction, however, these methods are currently limited by their high run-time and compute requirements -- a major limitation, if resolved, will only increase the use and effectiveness of this work.}
%

\bibliographystyle{ieeetr}   % or unsrt / plain / IEEEtran
\bibliography{references}

% \end{thebibliography}
\end{document}